\documentclass{article}
\usepackage{nips13submit_e,times}
\usepackage[bookmarks=false]{hyperref}
\usepackage{url}
\usepackage{amsmath}
\usepackage{amssymb}
\usepackage[labelfont=bf,font={footnotesize},singlelinecheck=off,justification=centering,aboveskip=0pt,belowskip=-10pt]{subcaption}
\usepackage[labelfont=bf,font={small},singlelinecheck=off,justification=raggedright,aboveskip=0pt]{caption}
\usepackage{graphicx}

\title{Deep Autoencoders for Dimensionality Reduction of High-Content Screening Data}

\author{
Lee Zamparo\thanks{\url{http://www.cs.toronto.edu/\~zamparo/}} \\
Department of Computer Science\\
University of Toronto\\
Toronto, ON, Canada \\
\texttt{zamparo@cs.toronto.edu} \\
\And
Zhaolei Zhang \\
Banting and Best Department of Medical Research \\
University of Toronto \\
Toronto, ON, Canada \\
\texttt{zhaolei.zhang@utoronto.ca} \\
}

\nipsfinalcopy 

\begin{document}

\maketitle

\begin{abstract}High-content screening uses large collections of unlabeled cell image data to reason about genetics or cell biology.  Two important tasks are to identify those cells which bear interesting phenotypes, and to identify sub-populations enriched for these phenotypes.  This exploratory data analysis usually involves dimensionality reduction followed by clustering, in the hope that clusters represent a phenotype.  We propose the use of stacked de-noising auto-encoders to perform dimensionality reduction for high-content screening.  We demonstrate the superior performance of our approach over PCA, Local Linear Embedding, Kernel PCA and Isomap.  
\end{abstract} 

\section{Introduction}
The use of machine learning methods to apply phenotype labels to cells has proven an effective way to uncover novel roles of genes in model organisms \cite{Vizeacoumar2010} as well as in human \cite{Fuchs2010b}. In an unsupervised model, characterizing all cells split across distinct populations presents a clustering problem over many millions of high dimensional data points. This complicates the clustering problem, and it is preferable to transform the data into a much lower dimensional space.  How to best perform the dimensionality reduction is far from clear. Particularly important qualities in this use case are the ability to scale to millions of data points, and the flexibility to model non-linear relationships between covariates in the map from higher to lower dimensional space.  Standard algorithms for dimensionality reduction fail in one or both of these criteria.  
\begin{itemize}
  \item PCA is fast once the covariance matrix is formed, but is not able to model any non-linear interactions.
  \item Kernel PCA \cite{Ham2004} can model more flexible relationships, but is impractical for large data sets due to the growth of the kernel matrix.
  \item Local Linear Embedding (LLE) \cite{Roweis2000b} and Isomap \cite{DeSilva2003} are impractical for large data sets as they require an large matrix decomposition that scales cubically in the number of data points.
\end{itemize}
Neural networks structured as auto-encoders (\cite{Hinton2006}) can model non-linear interactions, and scale well to large data sets.  They can also be trained with unlabeled data, which in the case of high-content screening is plentiful.  We investigated how well a class of auto-encoders (\cite{Vincent2008}) could be composed to create an efficient and flexible dimensionality reduction algorithm.

\section{Stacked De-noising Autoencoders}\label{sec:methods} 
The idea of composing simpler models in layers to form more complex ones has been successful with a variety of basis models, stacked de-noising autoencoders (abbrv. SdA) being one example \cite{Hinton2006,Ranzato2008a,vincent2010stacked}.  These models were frequently employed as \textit{unsupervised pre-training}; a layer-wise scheme for initializing the parameters of a multi-layer perceptron, which is subsequently trained by minimizing an appropriate loss function over real versus model-predicted labels of the data.  Where labeled data is plentiful, supervised training now reigns supreme, with pre-training overlooked in favour of randomly initialized models coupled with powerful new regularization techniques \cite{NitishMasters,Goodfellow2013}.  These methods are a poor fit for high-content screening, where biological expertise and time-intensive sampling makes labels scarce.  Unlabeled data remains plentiful, so pre-training is well-suited to this scenario.  We chose de-noising autoencoders since they are simple to train without incurring any sacrifice in competitive performance \cite{vincent2010stacked}.

\subsection{De-noising Autoencoders}
\begin{figure}[h]
\begin{center}

\fbox{\rule[-.5cm]{0cm}{4cm} \includegraphics[width=0.7\textwidth]{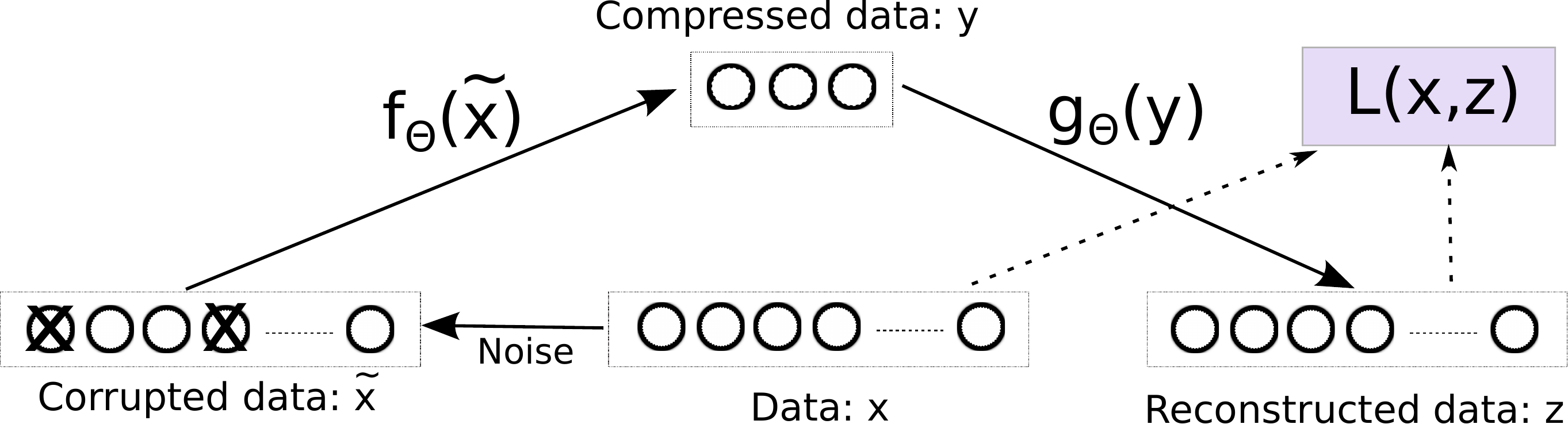} \rule[-.5cm]{0cm}{0cm}}
\end{center}
\caption{Schematic representation of a de-noising autoencoder (abbrv. dA).  The Stacked de-noising autoencoder model is a product of layering dA models such that the hidden layer of one model is treated as the input layer of the next dA.  The hidden layer of the top-most dA represents the reproduced data. }
\label{fig:sda_schematic}
\end{figure}

The de-noising autoencoder \cite{Vincent2008} takes input data $x \in \Re^{d}$, and maps it to a hidden representation $y \in [0, 1]^{n},\;n\ll d$ through a corrupted noisy mapping $y =  f_{\theta}(\tilde{x}) = sigmoid(W\tilde{x} + b)$, parametrized by $\theta = \left\lbrace W, b \right\rbrace $. $W$ is a $n \times d$ weight matrix and $b$ is a bias vector. The input $x$ is corrupted into $\tilde{x}$ by randomly setting a certain proportion of the values of $x$ to zero.  The latent representation $y$ is then mapped back to a reconstructed vector $z \in \Re^{d}$ via $z = g_{\theta'}(y) = W'y + b'$ with $\theta' = \left\lbrace W' , b' \right\rbrace$.  The parameters $W,d$ are set to minimize the reconstruction loss $L(x,z)$.  See the schematic in figure \ref{fig:sda_schematic}      

\subsection{Training}
To find the best architecture of layer sizes, we performed another grid search over both layer sizes as well as layers.  Hinton and Salakhutdinov \cite{Hinton2006} used a four layer auto-encoder for MNIST, but we began with 3 layer networks and tried both 4 and 5 layer networks.  Both 3 and 4 layer networks achieved better results measured by reconstruction error on a validation set.  For a given number of layers, each candidate model was a point on a grid with a step size of 100: ${700\dots1000} \times {500\dots900} \times {100\dots400} \times {50\dots10}$.  Each dA layer in every model was pre-trained using mini-batch stochastic gradient descent for 50 epochs, in batches of 100 samples.  The minimum mean reconstruction error for each layer was recorded.  After selecting the top 5 performing models for both 3 and 4 layer networks, we performed grid searches over each of the tunable hyper-parameters momentum, noise rate, learning rate, and weight decay.  We used Adagrad to adjust the learning rate appropriately for each dimension \cite{Duchi}.  Tuning the initial value of the base learning rate had the largest effect on performance. The other parameters showed much smaller effects.  Both the architecture and hyper-parameter searches were performed in parallel on a cluster with GPU capable nodes. All GPU code was written in python using Theano \cite{bergstra-theano-scipy}.

\section{Experiments}
We used data derived from images of yeast cell strains transformed to express a GFP fusion protein acting as a marker for DNA damage foci.  Each well in a 384 well plate contained an homozygous population of yeast bearing some set of genetic deletions.  Four field of view images per well were captured, each containing several hundred cells.  Post capture, the images were run through an image processing pipeline using CellProfiler, to segment the cells in each field, and represent each as a vector of intensity, shape and texture features \cite{Kamentsky2011a}.  The cell phenotypes in this study consisted of three classes: dna-damage foci, non-round nuclei, or wild-type\footnote{Yeast cells that displayed neither DNA-damage foci nor a non-round nucleus phenotype}.  A validation set of approximately 10000 labeled cells was generated by first manually labeling images for each phenotype, and then training an SVM in a one vs all manner to label the remaining cells.  The predicted labels were manually validated.  

\begin{figure}[h]
\begin{center}
\fbox{\rule[-.5cm]{0cm}{4cm} \includegraphics[scale=0.55]{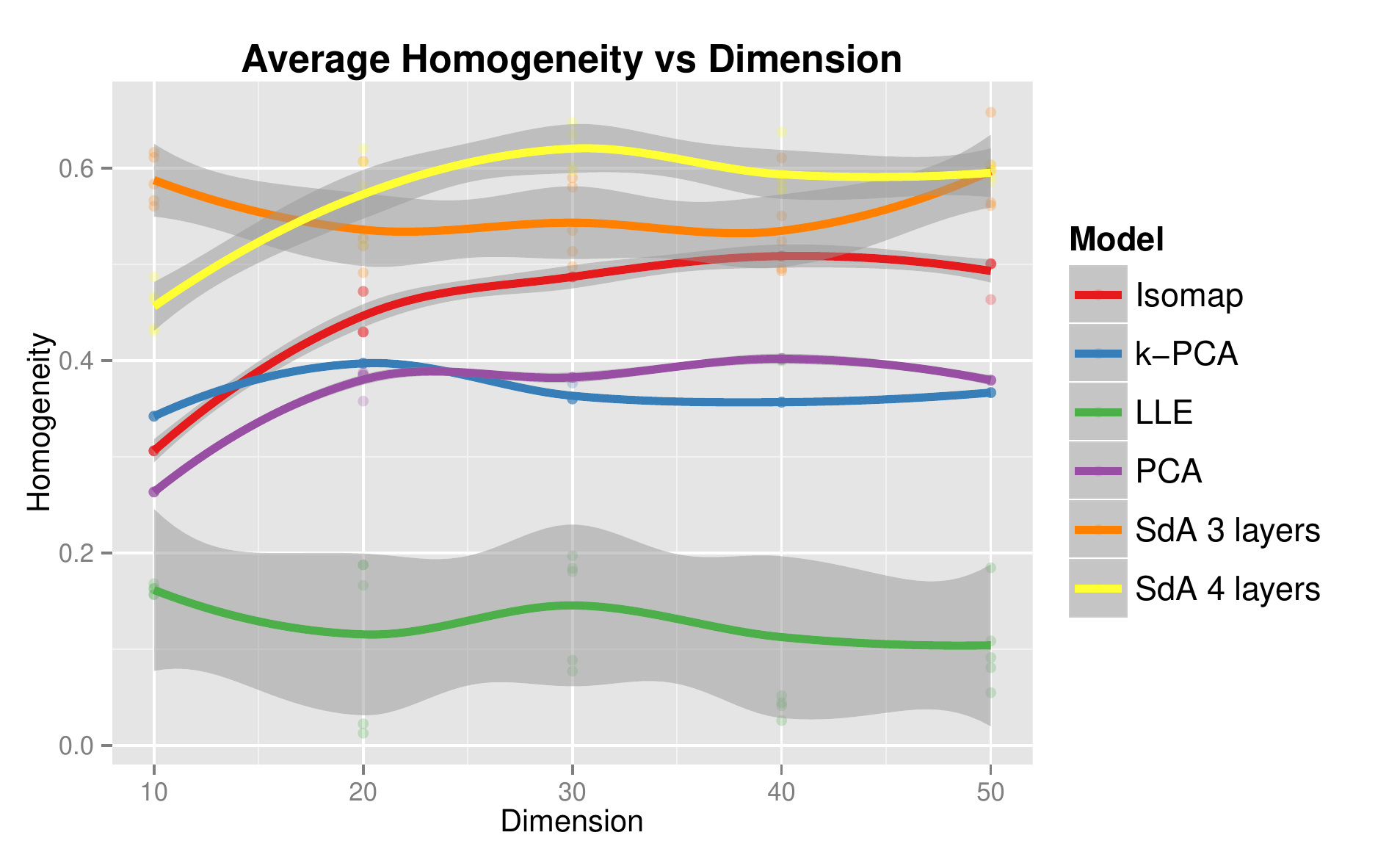} \rule[-.5cm]{0cm}{0cm}}
\end{center}
\caption{Homogeneity test results of 3, 4 layer SdA models versus comparators.  For each model, the data was reduced from 916 dimensions down to {50..10} dimensions.  For each model and embedding of the data, a Gaussian Mixture Model was run with randomly initialized parameters for 10 iterations.  Each of these runs was repeated 10 times. The parameters from the best performing run were used to initialize a GMM which was run until convergence, and the homogeneity was measured and recorded.  This process was repeated five times for each model, resulting in five homogeneity measurements. We report the average homogeneity (solid line) as well as the standard deviation (shadowed gray).}
\label{fig:homogeneity}
\end{figure}

The data for the validation set was scaled to have mean 0 and variance 1.  The scaled data was reduced using one of the candidate algorithms, followed by Gaussian mixture clustering using scikit-learn \cite{scikit-learn}.  The number of mixture components was chosen as the number of labels in the validation set.  All hyperparameters were chosen by cross-validation.  For each model and embedding of the data, a Gaussian Mixture Model was run with randomly initialized parameters for 10 iterations.  Each of these runs was repeated 10 times. The parameters from the best performing run were used to initialize a GMM which was run until convergence, and the homogeneity was measured and recorded.  This process was repeated five times for each model, resulting in five homogeneity measurements.  We report the average homogeneity (fit by loess) as well as the standard deviation \ref{fig:homogeneity}.

\section{Discussion}
\begin{figure}[h]
\begin{center}
\fbox{\rule[-.5cm]{0cm}{4cm} \begin{subfigure}[b]{0.49\textwidth}  
  \includegraphics[width=\textwidth]{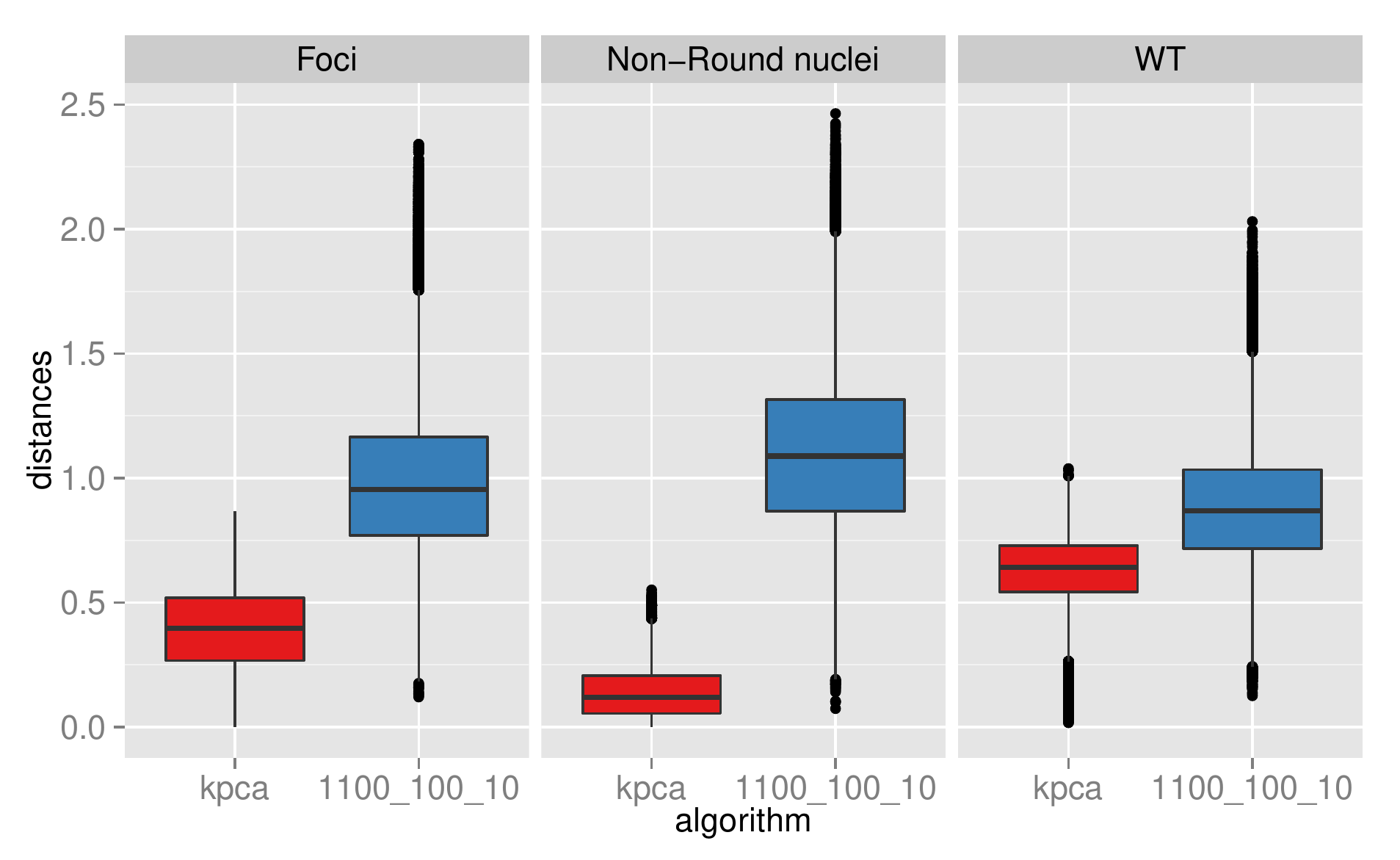}
  \subcaption{Within-class distance distributions, for SdA (denoted by its architecture) versus kernel PCA}
  \label{fig:intra_kpca}
\end{subfigure}%
\vspace{1cm}
\begin{subfigure}[b]{0.49\textwidth}
  \includegraphics[width=\textwidth]{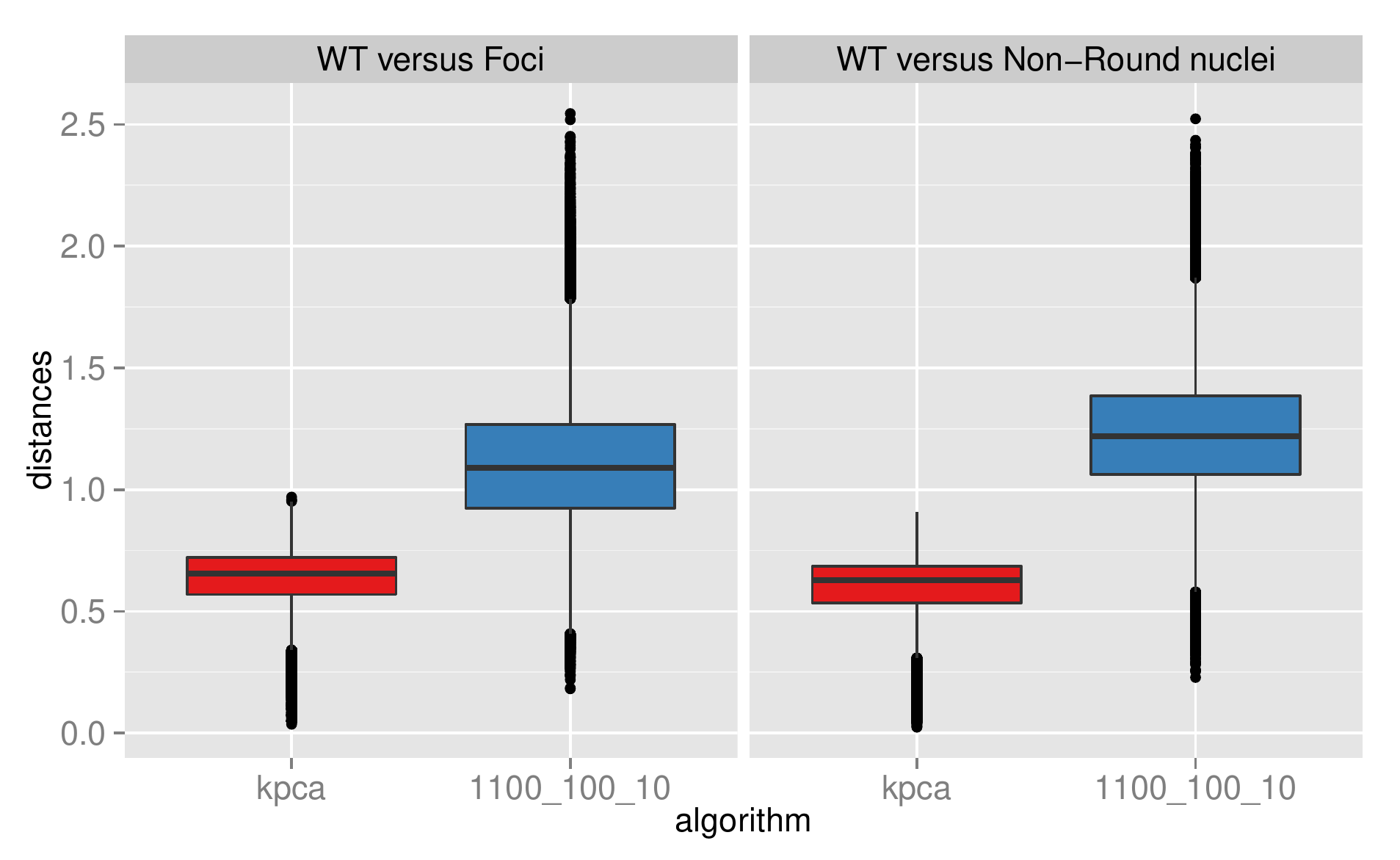}
  \subcaption{Distance distributions from points sampled from different classes, for SdA versus kernel PCA}
  \label{fig:inter_kpca}
\end{subfigure} \rule[-.5cm]{0cm}{0cm}}
\end{center}
\caption{Sampling from distributions over distances of data points reduced to 10 dimensions by both SdA, and kernel PCA.  While the SdA models do not always produce the tightest packing of points within classes (see left), they do consistently assign the widest mean distances to inter-class points.}
\label{fig:distances}
\end{figure}

Examination of figure \ref{fig:homogeneity} shows that SdA models were consistently ranked as the best models for dimensionality reduction in preparation for phenotype based clustering.  There is a loss of homogeneity incurred by using an algorithm that cannot learn non-linear transformations, as shown in the gap between Isomap and PCA. To try and understand what accounts for the gap between SdA models and the other algorithms, we randomly sampled output from each algorithm and computed estimates of the distribution over both the inter-label and intra-label distances between points.  Shown in figure \ref{fig:distances} is an example of the estimated distance distributions between kernel PCA and a sample three layer SdA model.  While the points sampled from kernel PCA have a smaller intra-label mean distance\footnote{Averaged over all three class labels}, the points sampled from the SdA model have a larger inter-label distance\footnote{Only one comparison is shown for conciseness, but this pattern is consistent across the other algorithms}. Therefore, a clustering algorithm applied to SdA reduced data was more reliably able to find assignments that reflected the phenotype labels. 
  
We have introduced deep autoencoder models for dimensionality reduction of high content screening data.  Mini-batch stochastic gradient descent allowed us to train using millions of data points, and the nature of the model allowed us to apply the resulting models to unseen data, circumventing the limitations of other comparable dimensionality reduction algorithms.  We also demonstrated that SdA models produced output that was more easily assigned to clusters that reflected biologically meaningful phenotypes.

\bibliographystyle{apalike}
\bibliography{mlcb2014}

\begin{thebibliography}{}

\bibitem[Bergstra et~al., 2010]{bergstra-theano-scipy}
Bergstra, J., Breuleux, O., Bastien, F., Lamblin, P., Pascanu, R., Desjardins,
  G., Turian, J., Warde-Farley, D., and Bengio, Y. (2010).
\newblock Theano: a {CPU} and {GPU} math expression compiler.
\newblock In {\em Proceedings of the Python for Scientific Computing Conference
  ({SciPy})}.
\newblock Oral Presentation.

\bibitem[{De Silva} et~al., 2003]{DeSilva2003}
{De Silva}, V., Tenenbaum, J. J.~B., and Silva, V.~D. (2003).
\newblock {Global versus local methods in nonlinear dimensionality reduction}.
\newblock {\em Advances in Neural Information Processing Systems 15}, 15(Figure
  2):705--712.

\bibitem[Duchi et~al., 2010]{Duchi}
Duchi, J., Hazan, E., and Singer, Y. (2010).
\newblock {Adaptive Subgradient Methods for Online Learning and Stochastic
  Optimization}.
\newblock Technical report, University of California at Berkeley, Berkeley.

\bibitem[Fuchs et~al., 2010]{Fuchs2010b}
Fuchs, F., Pau, G., Kranz, D., Sklyar, O., Budjan, C., Steinbrink, S., Horn,
  T., Pedal, A., Huber, W., and Boutros, M. (2010).
\newblock {Clustering phenotype populations by genome-wide RNAi and
  multiparametric imaging}.
\newblock {\em Mol Syst Biol}, 6(370).

\bibitem[Goodfellow et~al., 2013]{Goodfellow2013}
Goodfellow, I.~J., Warde-Farley, D., Mirza, M., Courville, A., and Bengio, Y.
  (2013).
\newblock {Maxout Networks}.

\bibitem[Ham et~al., 2004]{Ham2004}
Ham, J., Lee, D.~D., Mika, S., and Sch\"{o}lkopf, B. (2004).
\newblock {A kernel view of the dimensionality reduction of manifolds}.
\newblock In {\em Twenty-first international conference on Machine learning -
  ICML '04}, page~47, New York, New York, USA. ACM Press.

\bibitem[Hinton and Salakhutdinov, 2006]{Hinton2006}
Hinton, G.~E. and Salakhutdinov, R.~R. (2006).
\newblock {Reducing the dimensionality of data with neural networks.}
\newblock {\em Science (New York, N.Y.)}, 313(5786):504--7.

\bibitem[Kamentsky et~al., 2011]{Kamentsky2011a}
Kamentsky, L., Jones, T.~R., Fraser, A., Bray, M.-A., Logan, D.~J., Madden,
  K.~L., Ljosa, V., Rueden, C., Eliceiri, K.~W., and Carpenter, A.~E. (2011).
\newblock {Improved structure, function and compatibility for CellProfiler:
  modular high-throughput image analysis software.}
\newblock {\em Bioinformatics (Oxford, England)}, 27(8):1179--80.

\bibitem[Pedregosa et~al., 2011]{scikit-learn}
Pedregosa, F., Varoquaux, G., Gramfort, A., Michel, V., Thirion, B., Grisel,
  O., Blondel, M., Prettenhofer, P., Weiss, R., Dubourg, V., Vanderplas, J.,
  Passos, A., Cournapeau, D., Brucher, M., Perrot, M., and Duchesnay, E.
  (2011).
\newblock {Scikit-learn: Machine Learning in Python }.
\newblock {\em Journal of Machine Learning Research}, 12:2825--2830.

\bibitem[Ranzato et~al., 2008]{Ranzato2008a}
Ranzato, M., Boureau, Y.-l., and Cun, Y.~L. (2008).
\newblock {Sparse Feature Learning for Deep Belief Networks}.
\newblock In {\em Advances in Neural Information Processing Systems}, pages
  1185--1192.

\bibitem[Roweis, 2000]{Roweis2000b}
Roweis, S.~T. (2000).
\newblock {Nonlinear Dimensionality Reduction by Locally Linear Embedding}.
\newblock {\em Science}, 290(5500):2323--2326.

\bibitem[Srivastava, 2013]{NitishMasters}
Srivastava, N. (2013).
\newblock {\em {Improving Neural Networks with Dropout}}.
\newblock PhD thesis, Toronto.

\bibitem[Vincent et~al., 2008]{Vincent2008}
Vincent, P., Larochelle, H., Bengio, Y., and Manzagol, P.-A. (2008).
\newblock {Extracting and composing robust features with denoising
  autoencoders}.
\newblock In {\em Proceedings of the 25th international conference on Machine
  learning - ICML '08}, pages 1096--1103, New York, New York, USA. ACM Press.

\bibitem[Vincent et~al., 2010]{vincent2010stacked}
Vincent, P., Larochelle, H., Lajoie, I., Bengio, Y., and Manzagol, P.-A.
  (2010).
\newblock {Stacked denoising autoencoders: Learning useful representations in a
  deep network with a local denoising criterion}.
\newblock {\em The Journal of Machine Learning Research}, 11:3371--3408.

\bibitem[Vizeacoumar et~al., 2010]{Vizeacoumar2010}
Vizeacoumar, F.~J., van Dyk, N., {S Vizeacoumar}, F., Cheung, V., Li, J.,
  Sydorskyy, Y., Case, N., Li, Z., Datti, A., Nislow, C., Raught, B., Zhang,
  Z., Frey, B., Bloom, K., Boone, C., and Andrews, B.~J. (2010).
\newblock {Integrating high-throughput genetic interaction mapping and
  high-content screening to explore yeast spindle morphogenesis.}
\newblock {\em The Journal of cell biology}, 188(1):69--81.

\end{thebibliography}

\end{document}